\documentclass{article}

\pdfoutput=1

\usepackage{placeins}

\usepackage{amsmath,amsfonts,bm}
\usepackage{mathtools}
\usepackage{soul}

\newcommand{\ignore}[1]{}

\newcommand{\captionn}[1]{{\em (#1)}}

\DeclarePairedDelimiter\abs{\lvert}{\rvert}%
\DeclarePairedDelimiter\norm{\lVert}{\rVert}%

\makeatletter
\let\oldabs\abs
\def\abs{\@ifstar{\oldabs}{\oldabs*}}
\let\oldnorm\norm
\def\norm{\@ifstar{\oldnorm}{\oldnorm*}}
\makeatother

\newcommand{\figlabel}[1]{\label{Fi:#1}}
\newcommand{\tablabel}[1]{\label{Ta:#1}}
\newcommand{\seclabel}[1]{\label{Sc:#1}}

\def\Figref#1{Figure~\ref{Fi:#1}}

\def\Tabref#1{Table~\ref{Ta:#1}}

\def\secref#1{section~\ref{Sc:#1}}

\def\Secref#1{Section~\ref{Sc:#1}}

\def\Appref#1{Appendix~\ref{#1}}

\def\eqref#1{equation~\ref{#1}}
\def\Eqref#1{Equation~\ref{#1}}
\def\Eqrefs#1#2{Equations~\ref{#1} and~\ref{#2}}

\def\1{\bm{1}}

\def\vdelta{{\bm{\delta}}}

\def\vh{{\bm{h}}}

\def\vp{{\bm{p}}}

\def\vr{{\bm{r}}}

\def\vx{{\bm{x}}}
\def\vy{{\bm{y}}}

\def\evmu{{\mu}}

\def\evsigma{{\sigma}}

\def\evp{{p}}

\def\evx{{x}}
\def\evy{{y}}

\DeclareMathAlphabet{\mathsfit}{\encodingdefault}{\sfdefault}{m}{sl}
\SetMathAlphabet{\mathsfit}{bold}{\encodingdefault}{\sfdefault}{bx}{n}

\def\sR{{\mathbb{R}}}

\newcommand{\E}{\mathbb{E}}

\DeclareMathOperator*{\argmin}{arg\,min}

\usepackage{microtype}
\usepackage{graphicx}
\usepackage{subfigure}
\usepackage{booktabs} 

\newcommand{\tra}[1]{\renewcommand{\arraystretch}{#1}}

\usepackage{hyperref}

\usepackage[accepted]{icml2020}

\usepackage{multirow}
\usepackage{amsmath}
\usepackage{amssymb}
\usepackage{amsthm}
\usepackage{stmaryrd}
\usepackage{graphicx}
\usepackage{mathtools}
\usepackage{booktabs}
\usepackage{bbm}

\graphicspath{{figures/}}

\newtheorem{thm}{Theorem}[section] 
\newcommand{\thistheoremname}{}
\newtheorem{genericthm}[thm]{\thistheoremname}
\newenvironment{namedthm}[1]
  {\renewcommand{\thistheoremname}{#1}%
   \begin{genericthm}}
  {\end{genericthm}}
  
\newtheorem*{ngenericthm}{\thistheoremname}
\newenvironment{nnamedthm}[1]
  {\renewcommand{\thistheoremname}{#1}%
   \begin{ngenericthm}}
  {\end{ngenericthm}}

\DeclareMathOperator*{\obj}{obj}

\newcommand{\scode}[1]{\ensuremath{\mathtt{#1}}}

\newcommand\QYD{q[\vy|\vx+\vdelta, z] }

\newcommand\EQD{\E_{\QYD}[\chi(\vy)] }

\newcommand\sandp{S\&P}

\def\permille{\ensuremath{{}^\text{o}\mkern-5mu/\mkern-3mu_\text{oo}}}

\begin{document}
\twocolumn[

\icmltitle{Adversarial Attacks on Probabilistic Autoregressive Forecasting Models}

\begin{icmlauthorlist}
\icmlauthor{Rapha\"el Dang-Nhu}{to}
\icmlauthor{Gagandeep Singh}{to}
\icmlauthor{Pavol Bielik}{to}
\icmlauthor{Martin Vechev}{to}
\end{icmlauthorlist}

\icmlaffiliation{to}{Department of Computer Science, ETH Z\"urich, Switzerland}
\icmlcorrespondingauthor{Rapha\"el Dang-Nhu}{dangnhur@student.ethz.ch}
\icmlkeywords{Machine Learning, ICML}

\vskip 0.3in
]

\printAffiliationsAndNotice{} 

\begin{abstract}
We develop an effective generation of \emph{adversarial attacks} on neural models that output a sequence of probability distributions rather than a sequence of single values. This setting includes the recently proposed deep probabilistic autoregressive forecasting models that estimate the probability distribution of a time series given its past and achieve state-of-the-art results in a diverse set of application domains. The key technical challenge we address is effectively differentiating through the Monte-Carlo estimation of statistics of the joint distribution of the output sequence. Additionally, we extend prior work on probabilistic forecasting to the Bayesian setting which allows conditioning on future observations, instead of only on past observations. We demonstrate that our approach can successfully generate attacks with small input perturbations in two challenging tasks where robust decision making is crucial -- stock market trading and prediction of electricity consumption.
\end{abstract}

\section{Introduction}
Deep probabilistic autoregressive models have been recently integrated into the Amazon SageMaker toolkit and successfully applied to various kinds of sequential data such as handwriting \cite{graves2013generating}, speech and music \cite{oord2016wavenet}, images
\cite{van2016pixel,van2016conditional} and time series from a number of different domains~\cite{salinas2019deepar}.
At a~high level, given a~sequence of input values (i.e., pen-tip locations, raw audio signals or stock market prices), the goal is to train a~generative model that outputs an accurate sequence of next values, conditioned on all the previous values.
The main benefit of such probabilistic models is that they model the joint distribution of output values, rather than predicting only a~single best realization (i.e., the most likely value at each step).
Predicting a~density rather than just a single best value has several advantages -- it naturally fits the inherently stochastic nature of many processes, allows to assess the uncertainty and the associated risk, and has been shown to produce better overall prediction accuracy when used in forecasting tasks~\cite{salinas2019deepar}.

\paragraph{This Work}
We develop an efficient approach for generating \emph{adversarial attacks} on deep probabilistic autoregressive models.
The issue of adversarial robustness and attacks~\cite{szegedy2013intriguing, goodfellow2014explaining}, i.e., generating small input perturbations that lead to mis-predictions, is an important problem with large body of recent work.
Yet, to our best knowledge this is the first work that explores adversarial attacks in a new challenging setting where the neural network output is a sequence of probability distributions.
The difficulty in generating adversarial attacks in this setting is that it requires computing the gradient of an expectation that it is too complex to be analytically integrated~\cite{schittenkopf2000forecasting,salinas2019deepar} and is approximated using Monte-Carlo methods.

\paragraph{Differentiating through Monte-Carlo estimation of the Expectation}
We address the key technical challenge of efficiently differentiating through the Monte-Carlo estimation, a necessary part of generating white-box gradient based adversarial attacks, by using two techniques to approximate the gradient of the expectation. The first approach is the score-function estimator~\cite{glynn1990likelihood,kleijnen1996optimization} obtained by inverting the gradient and the expectation's integral. The second technique differentiates individual samples using a random variate \emph{reparametrizatrion} and originates from the variational inference literature~\cite{salimans2013fixed,kingma2013auto,rezende2014stochastic}.

\textbf{Main Contributions} 
We present the first approach for generating adversarial attacks on deep probabilistic autoregressive models by applying two techniques that differentiate through Monte-Carlo estimation of an expectation.
We show that the reparametrization estimator is efficient at generating adversarial attacks and outperforms the score-function estimator by evaluating on two domains that benefit from stochastic sequential reasoning -- stock market trading and electricity consumption.
We make our code, datasets and scripts to reproduce our experiments available online\footnote{\url{https://github.com/eth-sri/probabilistic-forecasts-attacks}}.
%

%
%

\section{Probabilistic Forecasting Models}\seclabel{background}

In this section, we formally describe the probabilistic autoregressive model used in prior works and throughout this paper. 
Most notably, the model described in \Secref{prelims} is based on the recent work of \scode{DeepAR}~\cite{salinas2019deepar}, which is now an inherent part of the Amazon SageMaker toolkit.
Further, in \Secref{bayesian} we describe an extension of this model to the Bayesian setting proposed in our work.

\subsection{Sequence Modeling: Preliminaries}\seclabel{prelims}

Given a sequential process $\vp = (p_t)_{1 \leq t_0 \leq T}$ and an index~$t_0$, we consider the task of modeling the distribution of predicted (future) values $\vp_{t_0:T} = (\evp_{t_0},\ldots,\evp_{T})$ given the observed (past) values $\vp_{1:t_0-1} = (\evp_1,\ldots,\evp_{t_0-1})$.
Using the chain rule, the joint distribution of predicted values conditioned on observed values $\Pr [\vp_{t_0:T} | \vp_{1:t_0-1}]$ can be written as a product of conditional distributions:
\begin{equation}
	\Pr[\vp_{t_0:T} | \vp_{1:t_0 - 1}] = \prod_{i = t_0}^{T} \Pr[\evp_i| \vp_{1:i-1}]
\end{equation}
In deep autoregressive models, a neural network is used to approximate the conditional distribution $\Pr[\evp_i | \vp_{1:i-1}]$ by a~parametric distribution $q_\theta[\evp_i | \vp_{1:i-1}]$ specified by learnable parameters $\theta$. This yields a joint model:

\begin{equation}
	q_\theta[\vp_{t_0:T} | \vp_{1:t_0-1}] = \prod_{i = t_0}^{T} q_\theta[\evp_i | \vp_{1:i-1}]
\end{equation}

This decomposed form for the joint distribution is general and independent of the particular neural architecture chosen for $q_\theta$.
In principle, any type of sequential model can be used including well-known architectures such as LSTMs~\cite{hochreiter1997long}, Temporal Convolutional Networks~\cite{bai2018empirical} or Transformers~\cite{vaswani2017attention}. Next, we describe a LSTM based instantiation of probabilistic autoregressive models.

\paragraph{Probabilistic Autoregressive Models}
Let $h$ be a function implemented by a LSTM network.
Given $h$, we compute the hidden state $\vh_i = h(\vh_{i-1},p_{i-1},\theta)$ for each time-step, conditioned on the previous hidden state~$\vh_{i-1}$, previous input value $p_{i-1}$ and the network parameters~$\theta$.
Then, the hidden state $\vh_i$ is used to generate a set of parameters $\psi(\vh_i)$ that specify a distribution with density $\ell_{\psi(\vh_i)}(p_i)$, giving the following form for the conditional distribution:

\begin{equation}
	q_\theta[\vp_{t_0:T} | \vp_{1:t_0-1}] = \prod_{i = t_0}^{T}   \ell_{\psi(\vh_i)}(p_i )
\end{equation}

The main difference here compared to the non-probabilistic models is that the network predicts parameters of a distribution rather than a single value.
Commonly used distributions in prior works are Gaussian distribution for real-valued data or the negative binomial distribution for count data~\cite{salinas2019deepar}.
When using Gaussian distribution, $\psi(\vh_i)$ has two components $\psi(\vh_i)\!=\!(\evmu(\vh_i),\evsigma(\vh_i))$, that correspond to the mean and the standard deviation, respectively.
The density is defined as:
\begin{equation}
\ell_{\psi(\vh)}(p) = \frac{1}{\evsigma(\vh) \sqrt{2 \pi}}\exp \left[ - \frac{1}{2} \left( \frac{p - \evmu(\vh)}{\evsigma(\vh)} \right)^2 \right]
\end{equation}

Note, that the choice of a Gaussian distribution corresponds to the assumption that each value is normally distributed conditioned on past values -- a hypothesis which has to be assessed per application domain.
In what follows, to make a~clear distinction between the network inputs and outputs, we use $\vx = (\evx_1,\ldots,\evx_{t_0 - 1}) := \vp_{1:t_0 - 1} $ to denote the inputs (i.e., observed values) and $\vy = (\evy_1,\ldots,\evy_{T+1-t_0}) := \vp_{t_0:T}$ to denote the outputs (i.e., the predicted values).

\paragraph{Inference}
Performing inference for probabilistic autoregressive models corresponds to characterizing the joint distribution of the output sequence $\vy$. This includes
estimating $n$-steps ahead both, the mean and the standard deviation of the value $\evy_n$, via the first and second moments $\E_{q_\theta[\vy|\vx] }[\evy_{n}]$ and $\E_{q_\theta[\vy|\vx] }[\evy_{n}^2]$. More generally, given the space of output sequences $\mathcal{Y}$ and any statistic $\chi : \mathcal{Y} \rightarrow \mathbb{R}$, we consider the task of estimating the expectation $\E_{q_\theta[\vy|\vx] }[\chi(\vy)]$.
The main challenge here is the complexity of the underlying integral on the distribution $q_\theta[\vy|\vx]$, which is in general not analytically solvable.

During training, one can either use scheduled sampling~\cite{scheduled_sampling}, where a single sample from the distribution $\ell_{\psi(\vh_i)}(\cdot)$ is used, or avoid this issue completely by using teacher forcing~\cite{williams1989learning}, where the deterministic ground truth value for $y_i$ is fed back into the network in the next time-step.
This setting is solvable but only because the prediction is only for a single next step. 
However, when performing iterated prediction at test time, the value used in the feedback loop is sampled from the predicted distribution of $\evy_i \sim \ell_{\psi(\vh_i)}(\cdot)$. Therefore, the next hidden state $\vh_{i+1}$ depends on the randomness introduced in sampling $\evy_i$.
This yields an arbitrarily complex form for the joint distribution $q_\theta[\vy|\vx]$.
To address this issue, prior works perform Monte-Carlo inference~\cite{schittenkopf2000forecasting,salinas2019deepar} to approximate the expectation as:

\begin{equation}
\E_{q_\theta[\vy|\vx] }[\chi(\vy)] \simeq \frac{1}{L} \sum_{l = 1}^L \chi(\vy^l)
\end{equation}

That is, the Monte-Carlo estimations of the expected value of $\chi(\vy)$ is computed using $L$ generated samples $\vy^1,\ldots,\vy^L$ for the output sequence.

\subsection{Extension to Bayesian Setting}\seclabel{bayesian}
We extend the probabilistic autoregressive models presented in this section to the Bayesian setting, where the output sequence can be conditioned on arbitrary values (i.e., both past and from the future).
Formally, we define an observation function $ \gamma\colon \sR^m \rightarrow \{\scode{true}, \scode{false} \}$, which takes as input a
sequence of values and outputs a boolean denoting whether the observation holds.
As an example, using $\gamma(\vy) = (y_{10} \geq 3)$ denotes that we would like the model to predict values $y_{1:9}$ conditioned both on the inputs $\vx$ as well as on the observation $\evy_{10} \geq 3$.

There are several cases for which the Bayesian setting is useful: \captionn{i} some of the data is missing (e.g., due to sensor failures), \captionn{ii} to allow encoding prior beliefs about the future evolution of the process, or \captionn{iii} to evaluate complex domain-specific statistics. The financial domain offer good examples for (iii), in pricing exotic derivatives such as barrier options~\cite{rich1994mathematical} whose existence depends upon the underlying asset's price breaching a preset barrier level.

To remove clutter, in the remainder of the paper we will use $z = \gamma(\vy)$ to denote the output of the observation function when evaluated on $\vy$. In this Bayesian setting, the expectation $\E_{q_\theta[\vy|\vx, z] }[\chi(\vy)]$ can be estimated via Monte-Carlo importance sampling as:

\begin{equation}
\E_{q_\theta[\vy|\vx, z] }[\chi(\vy)] \simeq \frac{ \sum_{l = 1}^L \chi(\vy^l)  q_\theta[z|\vx, \vy^l]}{\sum_{l = 1}^L q_\theta[z|\vx, \vy^l]  }
\end{equation}

This corresponds to generating samples from the prior distribution $q_\theta[\vy|\vx]$, and reweighing with Bayes rules. Note, that this formula includes the former by using~$z=\scode{true}$.

\section{Adversarial Attacks on Probabilistic Forecasting Models}\seclabel{attack}
In this section, we present our approach for generating adversarial attacks on deep probabilistic forecasting models.
We start by formally defining the problem statement suitable for this setting and then we describe two practical adversarial attacks that address it.

\paragraph{Adversarial Examples}

Recall that in the canonical classification setting, adversarial examples are typically found by solving the following optimization problem~\cite{szegedy2013intriguing, papernot2016limitations,carlini2017towards}:
\begin{equation}\label{eq:adv_ex}
        \argmin_\vdelta  || \vdelta || \text{ s.t } f(\vx+\vdelta) = t
\end{equation}

where $f$ is a classifier, $\vx$ is an input (i.e., an image), $t$ is the desired adversarial output (the target), and $\vdelta$ is the minimal perturbation (according to a~given norm) applied to the input image such that the classifier $f$ predicts the desired output $t$.

To make the above formulation applicable to probabilistic forecasting models we perform two standard modifications -- \captionn{i} we replace the hard equality constraint with an easier to optimize soft constraint that captures the distance between real values, and \captionn{ii} we replace the single value output with the expected value of a given statistic of the output joint probability distribution.
Applying the first modification leads to the following formulation:

\begin{equation}\label{eq:adv_ex_2}
        \argmin_\vdelta \phi(f(\vx+\vdelta), t) \text{ s.t } ||\vdelta|| \leq \epsilon
\end{equation}
That is, we use a soft constraint that minimizes the distance between the target and the predicted value, subject to a~given tolerance $\epsilon$ on the perturbation norm.
Applying the second modification corresponds to replacing $f(\vx+\vdelta)$ with the expected value $\E_{q[\vy|\vx+\vdelta, z] }[\chi(\vy)]$, where $z$ is an observation over outputs as defined in \Secref{bayesian}, and $\chi :  \sR^m \rightarrow \sR$ is a statistic of the output sequence.
Overall, this leads to the following problem statement.

\paragraph{Problem Statement}
Let $f\colon \sR^n \rightarrow \mathcal{D}(\sR^m)$ be a function (i.e., a probabilistic neural network) that takes as input a~sequence of values $\vx \in \sR^n$ and outputs a probability distribution $f(\vx)$ with density $q[\vy|\vx]$ that can be sampled from to obtain a concrete output sequence $\vy \in \sR^m$. Given an observation variable $z$, a statistic $\chi :  \sR^m \rightarrow \sR$ of the network output distribution and a target $t$, the goal of the adversarial attack is to find a perturbation $\vdelta$ that solves the following optimization problem:
\begin{equation}\label{eq:problem_statement}
        \argmin_\vdelta \phi(\EQD, t) \text{ s.t } ||\vdelta|| \leq \epsilon
\end{equation}

\subsection{Practical Attack on Probabilistic Networks}\label{sec:our_attack}

The constrained minimization problem defined in \Eqrefs{eq:adv_ex_2}{eq:problem_statement} has repeatedly been identified as very difficult to solve in the adversarial attacks literature~\cite{szegedy2013intriguing,carlini2017towards}. As a result, we instead follow the approach of Szegedy et. al.~\citeyear{szegedy2013intriguing} and solve an adjusted optimization problem. Given a real hyper-parameter $c \in \sR^+$, we aim at minimizing:
\begin{equation}\label{eq:obj}
	\obj ( \vdelta ) :=||\vdelta|| + c \cdot \phi(\EQD, t)
\end{equation}
via gradient-descent. The attack is run with different values of $c$, and the final value is chosen to ensure that the hard constraint $||\vdelta|| \leq \epsilon$ is satisfied.

While optimizing the objective function in \Eqref{eq:problem_statement} is standard~\cite{szegedy2013intriguing, goodfellow2014explaining, Kurakin17, Kurakin17Physical}, the crucial aspect in ensuring efficient gradient descent is to obtain a good estimation of the objective function's gradient. In particular, this involves computing:
\[
	 \nabla_{\vdelta} \E_{q[\vy|\vx+\vdelta, z]}[\chi(\vy)]
\]
The difficulty of computing this gradient comes from the fact that the expectation can not be analytically computed, but only approximated via Monte-Carlo methods. Informally, it raises the question of how to efficiently differentiate through the Monte-Carlo estimation. We compare two different ways of performing this differentiation, described next.

\subsubsection{Score-function Estimator}\seclabel{score_function_estimator}

The first approach is to express the gradient of the expectation as an expectation over the distribution $q[\vy|\vx+\vdelta, z]$ by inverting the gradient and integral, and estimate the resulting expectation via Monte-Carlo methods. This technique is known under different names in the literature: score-function method~\cite{glynn1990likelihood}, \textsc{REINFORCE}~\cite{williams1992simple}, or log-derivative trick.
Below we show how this applies to our setting.

\begin{namedthm}{Score-function Estimator}\label{thm:score_function}
In the general Bayesian setting where $\vy \sim q[\cdot|\vx+\vdelta, z]$, the score-function gradient estimator of the expected value of $\chi(\vy)$ is:
\begin{align*}
	& \nabla_\vdelta \EQD \\
	\simeq &\frac{ \sum_{l = 1}^{L} \chi(\vy^l) q[z|\vx+\vdelta, \vy^l]  \nabla_\vdelta \log(q[\vy^l|\vx+\vdelta, z]) }{ \sum_{l = 1}^{L} q[z|\vx+\vdelta, \vy^l] }
\end{align*}
where $\vy^l$ is sampled from the prior distribution $q[\vy|\vx+\vdelta] $, and $q[z|\vx+\vdelta, \vy]$ denotes the probability that $z$ is true knowing that $\vy^l$ is generated.
\end{namedthm}

The proof of~\ref{thm:score_function} is given in the supplementary material. Note that in the non-Bayesian setting where the observation $z$ is always true, we have $q[z|\vx+\vdelta, \vy^l] = 1$ and we obtain a simpler form:
\begin{align*}
 \nabla_\vdelta \E_{q[\vy|\vx+\vdelta]}[ \chi(\vy)]
	\simeq \frac{ 1}{L}\sum_{l = 1}^{L}  \chi(\vy^l) \nabla_\vdelta \log(q[\vy^l|\vx+\vdelta]) 
\end{align*}

While this estimator allows for generating adversarial perturbations, we observe that it has two drawbacks -- high-variance and high sampling complexity.
\paragraph{High Variance} Score-function estimators typically lead to slow convergence because they suffer from high variance~\cite{ranganath2014black}. It is due to the fact that they operate in a black-box way with respect to the gradients of the network $f$ and the statistic $\chi$.

\paragraph{Complexity of the Bayesian Setting}
The score-function requires computing the gradient of $q[ \vy^l|\vx+\vdelta,z]$ with respect to $\vdelta$. This is always possible in the special case when the observation $z$ is constantly true, however in the general setting this might require another step of sampling, which makes the estimator overly complex.

\subsubsection{Reparametrization estimator}

The second estimator is based on the \emph{reparametrization~trick}. It reparametrizes the output distribution $\vy$ in terms of auxiliary random variables whose distribution does not depend on $\vx$, in order to make individual samples $\vy^l$ differentiable with respect to $\vdelta$. The differential $\partial \vy^l/\partial \vdelta$ has \emph{a priori} no specific meaning when the distribution from which $\vy^l$ is sampled depends on $\vdelta$. However, if $\vy^l \sim q[\cdot |\vx+\vdelta] $ can be reparametrized as $\vy^l = g_\vx(\vdelta,\bm{\eta})$, where $\bm{\eta}$ is a random variable whose distribution is independent from $\vdelta$, then it makes sense to define the differential of $\vy^l$ with respect to $\vdelta$ as $\partial \vy^l/\partial \vdelta = \partial g_\vx/\partial \vdelta$.

Reparametrization estimators were first proposed as a way of mitigating the variance problems of score-function estimators~\cite{salimans2013fixed,kingma2013auto,rezende2014stochastic}. However, to our best knowledge, they have not been used in a Bayesian setting where the estimator to differentiate uses importance sampling. 

\begin{namedthm}{Reparametrization Estimator}\label{thm:reparametrization_estimator}
Assume there exists a differentiable transformation $g_\vx(\vdelta,\bm{\eta})$ such that the random variable $\vy \sim q[\cdot |\vx+\vdelta] $ can be reparametrized as $\vy = g_\vx(\vdelta,\bm{\eta})$, where $\bm{\eta}$ is an independent random variable whose marginal distribution $p(\bm{\eta})$ is independent from $\vdelta$.  Then the importance sampling reparametrization estimator of the expectation's gradient is:
\begin{align*}
	 &\nabla_\vdelta \E_{q[\vy|\vx+\vdelta, z]}[ \chi(\vy)] \\
	 \simeq &\nabla_\vdelta \left( \frac{ \sum_{l = 1}^L \chi(g_\vx(\vdelta,\bm{\eta}^l))  q[z|\vx+\vdelta, g_\vx(\vdelta,\bm{\eta}^l)]}{ \sum_{l = 1}^L q[z|\vx+\vdelta, g_\vx(\vdelta,\bm{\eta}^l)]  } \right) \\ 
\end{align*}
where for $1 \leq l \leq L$, $\bm{\eta}^l$ is sampled from the distribution $p(\bm{\eta})$, and $\vy^l = g_\vx(\vdelta,\bm{\eta}^l)$.
\end{namedthm}

A proof of \ref{thm:reparametrization_estimator} is given in the supplementary material.

\subsection{Reparametrization of Probabilistic Networks}

Here, we discuss the question of reparametrizing probabilistic autoregressive models. The stochasticity of such architectures comes from the iterated sampling of $\evy_i\!\sim\!\ell_{\psi(\vh_i)}(\cdot)$. Assuming a Gaussian likelihood, the value $y_i$ follows a normal distribution $\mathcal{N}(\mu(\vh_i)|\sigma(\vh_i)^2)$. Let  $\bm{\eta} = (\eta_1,\ldots,\eta_{T-t_0})$ be a standard normal random vector, i.e., all of its components are independent and each is a zero-mean unit-variance normally distributed random variable. Iteratively writing:
\[
y_i \sim \mu(\vh_i) + \eta_i \cdot \sigma(\vh_i)
\]
for all $i$ such that $1\!\leq\!i\!\leq\!T - t_0$ yields a valid reparametrization. This simple reasoning applies to the particular implementation described in~\cite{salinas2019deepar} and adapts readily to any kind of likelihood parameterized by location and scale, such as Laplace or logistic distribution.

The case of mixture density likelihoods is more complex as they do not enter this category of "location-scale" distributions, and their inverse cumulative density function does not admit a simple closed form. The problem of how to adapt reparametrization to mixture densities is outside the scope of this paper, and we refer to the relevant literature~\cite{graves2016stochastic,figurnov2018implicit,jankowiak2018pathwise} for more information about this question.

\section{Case Study: Stock Market Trading}\label{application}

In this section, we apply the probabilistic autoregressive models and discuss the types of adversarial attacks in the domain of financial decision making.

\paragraph{Output Sequence Statistics}
While a given machine learning model is typically trained to predict the future stock prices given its past, various statistics of the output sequence are used in downstream algorithmic trading and option pricing tasks.
This is the reason why the approach presented so far already assumed presence of such statistics.
The different statistics used in our evaluation are shown in \Tabref{statistics} and include predicting cumulated stock return, pricing derivatives such as European call and put options~\cite{black1973pricing}, as well as an example of a~binary statistic that predicts the success probability of limit orders~\cite{handa1996limit}.
All statistics are defined with respect to the last known price, denoted as~$x_{-1}$.

\subsection{Probabilistic Autoregressive Models Performance}\label{sec:performance}
Before we show the effectiveness of generative adversarial attacks, we first demonstrate that using probabilistic autoregressive models leads to state-of-the-art results.
We use two baselines as the current state-of-the-art for financial predictions: LSTM networks~\cite{fischer2018deep}, and Temporal Convolutional Networks (TCN)~\cite{borovykh2017conditional}.
We provide detailed description of all the training hyper-parameters, the dataset used (S\&P 500) and extended version of all the experiments in the supplementary material.

\begin{table}
\caption{Financial gain on algorithmic trading tasks for different horizons $h$ and portfolio sizes $k$  (expressed per mille \permille). An extended version is included in the supplementary material.}\tablabel{trading}
\tra{1.3}
\center
\addtolength{\tabcolsep}{-5pt}
\begin{tabular}{c c c c c }
\toprule
  \multicolumn{2}{c}{Params} &  \multicolumn{2}{c}{Non-probabilistic} & Probabilistic \\ 
 \cline{1-2}\cline{3-5} 
 $h$ & $k$ &  \scode{TCN} & \scode{LSTM}  & \scode{LSTM} \\[-0.7em]
  & &  \scriptsize \cite{borovykh2017conditional} & \scriptsize \cite{fischer2018deep} & \scriptsize This Work\\
\midrule
   1 & 10 &   3.53 ($\pm$ 0.49) & \textbf{4.89 ($\pm$ 0.39)} & 4.37 ($\pm$ 0.51) \\
   1 & 30 &   1.74 ($\pm$ 0.30) & \textbf{2.41 ($\pm$ 0.24)} & 2.35 ($\pm$ 0.23) \\
  1 & 100 &  0.70 ($\pm$ 0.19) & 0.93 ($\pm$ 0.1) & \textbf{0.99 ($\pm$ 0.12)} \\
  \midrule
  5 & 10 &  5.57 ($\pm$ 1.93) & 8.86 ($\pm$ 1.03) & \textbf{9.02 ($\pm$ 1.52)} \\
  5 & 30 &  3.40 ($\pm$ 1.36) & 5.34 ($\pm$ 0.61) & \textbf{5.66 ($\pm$ 0.87)} \\
  5 & 100 &  1.64 ($\pm$ 0.78) & 2.47 ($\pm$ 0.28) & \textbf{2.70 ($\pm$ 0.48)} \\
  \midrule
  10 & 10 &  6.21 ($\pm$ 3.52) & \textbf{9.68 ($\pm$ 1.58)} & 9.55 ($\pm$ 2.30) \\
  10 & 30 &  4.28 ($\pm$ 2.69) & 6.39 ($\pm$ 0.76) & \textbf{6.63 ($\pm$ 1.70)} \\
  10 & 100 &  2.09 ($\pm$ 1.58) & 3.12 ($\pm$ 0.52) & \textbf{3.48 ($\pm$ 1.03)} \\
\bottomrule
\end{tabular}
\end{table}
\begin{table*}[t]
\caption{Definition of various output sequence statistics used in our work (left) and performance of various models used to predict them (right). $h$ is the prediction horizon and $\pi$ the price of the option. The comparison metric is Ranked Probability Skill~\cite{weigel2007discrete} of the prediction (lower scores correspond to better predictions). An extended version is provided in the supplementary material.}\tablabel{statistics}
\tra{1.3}
\center
\vspace{-0.4em}
\begin{tabular}{l c c c c c c c c}
\toprule
\multicolumn{2}{c}{Statistics} & & \multicolumn{2}{c}{Params} & & \multicolumn{2}{c}{Non-probabilistic} & Probabilistic \\ 
\cline{1-2} \cline{4-5}\cline{7-9} 
Name & $\chi(\vy)$ & & $h$ & $\pi$ & & \scode{TCN} & \scode{LSTM}  & \scode{LSTM} \\[-0.7em]
& & & & & & \scriptsize \cite{borovykh2017conditional} & \scriptsize \cite{fischer2018deep} & \scriptsize This Work\\
\midrule
Cum. Return 	& $y_{h}/x_{-1} - 1$ & & 10 & - & & 1.548 ($\pm$ 0.029) & 1.541 ($\pm$ 0.019) & \textbf{1.002 ($\pm$ 0.008)} \\
European Call 	& $\max(0,y_h / x_{-1} - \pi)$ & & 10 & 1 &  & 1.122 ($\pm$ 0.002) & 1.121 ($\pm$ 0.002) & \textbf{0.982 ($\pm$ 0.005)} \\
European Put 	& $\max(0,p-y_h / x_{-1})$ & & 10 & 1 & & 1.302 ($\pm$ 0.003) & 1.300 ($\pm$ 0.002) & \textbf{0.974 ($\pm$ 0.005)} \\
Limit Sell 		& $\mathbbm{1}\big[ \max(\vy_{1:h})/ x_{-1} \geq \pi \big]$ & & 10 & 1.05 & & 1.516 ($\pm$ 0.001) & 1.514 ($\pm$ 0.002) & \textbf{0.940 ($\pm$ 0.006)} \\
Limit Buy 		& $\mathbbm{1} \big[ \min(\vy_{1:h}) / x_{-1} \leq \pi \big]$ & & 10 & 0.95 & & 1.412 ($\pm$ 0.000) & 1.410 ($\pm$ 0.001) & \textbf{0.958 ($\pm$ 0.008)} \\
\bottomrule
\end{tabular}
\end{table*}

\paragraph{Long-Short Trading Strategies}\label{sec:trading}

Given a prediction horizon $h \in \llbracket 1, 10 \rrbracket$, we analyze the characteristics of the following portfolio: at time-step $t$, buy (long) the $k$ stocks for which the model predicts the highest gain, and sell (short) the $k$ stocks with the highest predicted loss.
This task is a~generalization of the one presented in~\cite{fischer2018deep}, where only direct prediction ($h = 1$) is considered.

Formally, we consider the cumulative return statistic $\chi_x(\vy) = \evy_h/x_{-1} - 1$ of the output sequence, which corresponds to the gain of investing one dollar in the stock at time $t$, and then selling at time $t + h$.
In a non-Bayesian setting, we estimate the expectation $\E_{q(\vy|\vx)}[\chi_x(\vy)]$ via Monte-Carlo sampling for each stock, and buy (or sell) the stocks for which the estimate is the highest (or the lowest). 
Note that this setting also applies to the deterministic baselines, it suffices to consider that $q(\vy|\vx)$ is a Dirac distribution centered in the deterministic prediction.

The performance of all models is summarized in \Tabref{trading}.
We can see that the TCN is consistently outperformed by both probabilistic and non-probabilistic LSTM models. For the probabilistic model, we observe that it is generally outperformed by the LSTM for direct prediction ($h = 1$), but it has better performance on iterated prediction ($h > 1$), provided that enough samples are used for Monte-Carlo estimation. We observe that a large number of samples (at least 1000) is required to match the LSTM performance.
We provide extended evaluation results in the supplementary material, including the effect of the number of samples.

\paragraph{Quality of the Probabilistic Forecast}
\Tabref{statistics} shows evaluation of the forecast quality for each of the statistics described earlier.
To compare deterministic and probabilistic forecast, we use as metric the Ranked Probability Skill (RPS)~\cite{weigel2007discrete} of the prediction.
However, because it applies only to predictions with finite output space, we first discretize the output before we apply RPS\footnote{There exists a continuous version~\cite{gneiting2007strictly}, but it is impractical for our setting because of the memory consumption of computing the score: we favor metrics computable in an on-line fashion with respect to the sampling process.}. Here, lower score means better prediction.
We provide extended evaluation results in the supplementary material, including multiple different values for the horizon $h$, price $\pi$ and the number of samples for Monte-Carlo estimation.

\subsection{Market Manipulations}\label{sec:attack_description}

The possibility of artificially influencing stock prices to make profit has always been a major problem of financial markets~\cite{allen1992stock, diaz2011analysis, ougut2009detecting}.
In our work, we focus on trade-based manipulation, in which a trader attempts to manipulate the price of a stock only by buying and then selling, without taking any other publicly observable action.
The core of such an attack is to anticipate the reactions of other agents to a provoked market event, in order to drive the price up or down.
In order to decrease the cost and visibility of the attack, an additional constraint for the manipulating trader is to minimize the amplitude of the perturbation.
This creates a natural connection with finding adversarial perturbations over the inputs~$\vx$.

\paragraph{Adversarial Attacks Scenario}
To measure the perturbation size, we choose a variant of the Euclidean norm specifically tailored to stock price data, defined in Equation~\ref{eq:norm}, where each component is normalized by the corresponding price $x_i$. This normalization aims at capturing the fact that stock prices are fixed for an arbitrary unit quantity of the underlying asset, and thus should be invariant with respect to multiplication by a scalar. Besides, we add a box constraint to the perturbed prices such that they remain positive. This constraint is enforced using projected gradient descent.
\begin{equation}\label{eq:norm}
\textstyle ||\vdelta||_{\vx} = \left(\sum_{i = 1}^{\abs{\vx}}  \left( \frac{\delta_i}{\evx_i} \right)^2\right)^{1/2}
\end{equation}

\section{Experimental Results}\label{eval}
In this section, we evaluate the effectiveness of our approach for generating adversarial attack on probabilistic autoregressive models.
The two key results of our evaluation are:
\begin{itemize}
\item The reparametrization estimator leads to significantly better adversarial examples (i.e., with smaller perturbation norm $\epsilon$) than the score-function estimator.
\item The reparametrization estimator successfully generates adversarial attacks for a number of different tasks. For example, using a small perturbation norm $\epsilon=0.016$\footnote{Here, $\epsilon=0.016$ corresponds to perturbing one price in the sequence by $1.6\%$, or 10 prices by $0.51\%$, or 100 prices by $0.16\%$.} the attack is powerful enough to cause financial loss when applied to stock market trading.

\end{itemize}

\begin{figure*}
  \includegraphics[width=\linewidth]{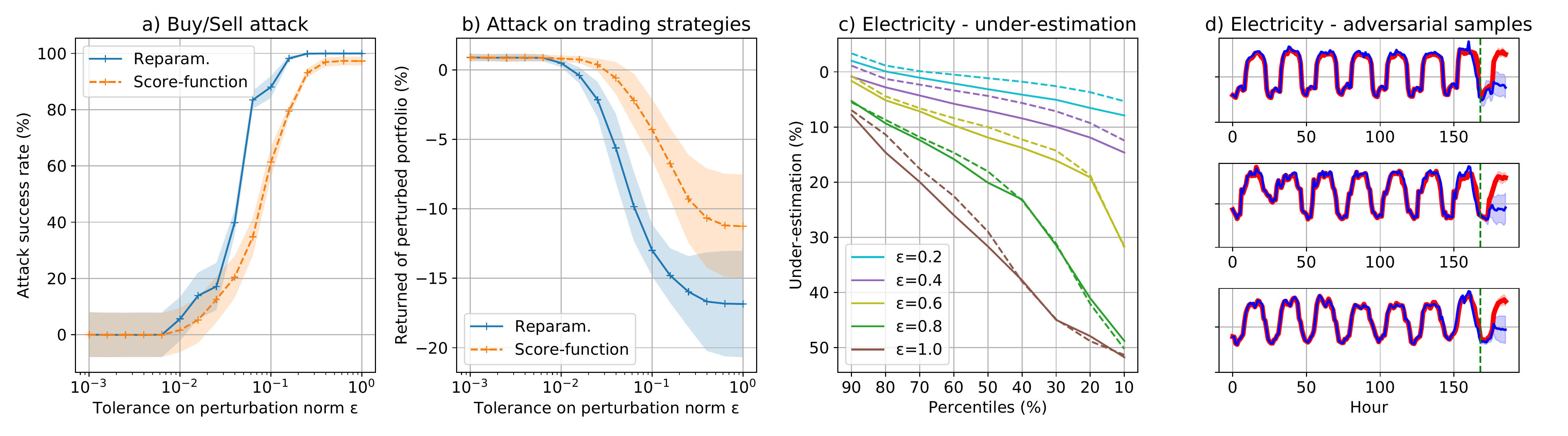}
  \vspace{-1.5em}
  \caption{  a) Success rate of the classification attack for different perturbation norms. b) Impact of the adversarial attack against trading strategies on financial gain, with portfolio size of $k = 10$. For both a) and b), standard deviation (shaded) is computed across test years c)~Under-estimation of electricity consumption for reparametrization (continuous) and score-function (dashed) estimators. For example, the graph shows that with $\epsilon = 1.0$, the attack using reparametrization estimator leads to under-estimation of at least $20\%$ (y-axis) for $70\%$ of samples (x-axis). d) Under-estimation adversarial samples for the electricity dataset, with $\epsilon = 0.9$. Red curve is the original sample, blue curve is the generated adversarial sample. The vertical dashed line separates the input sequence from the network's prediction.}
  \figlabel{attack}
\end{figure*}

\paragraph{Datasets} We evaluate on the following two datasets:

\textit{\sandp{} 500 dataset}, which contains historical prices of \sandp{} 500 constituents from 1990/01 to 2000/12.
We consider study periods of four consecutive years. 
The first three years serve as training data, while the last year is used for out-of-sample testing. 
The different periods have non-overlapping test years, resulting in eight different periods (for each we train four different models) with test year going from 1993 to 2000. 
We generate input-output samples by considering sequences of 251 consecutive daily prices for a~fixed constituent. 
The first 241 prices serve as input $\vx$, while the last 10 are the ground truth output $\vy$. 
We ensure that output sequences from the training and test sets do not overlap and reserve $\approx 15\%$ of training samples as a validation set.
We use the same preprocessing as in prior work (described in supplementary material) and train our own probabilistic autoregressive model (described in \Secref{background}). The order of magnitude of the cumulated test set size is $10^6$.

\textit{UCI electricity dataset}\footnote{\url{https://archive.ics.uci.edu/ml/datasets/ElectricityLoadDiagrams20112014}}, which contains the electricity consumption of 370 households from 2011 to 2014, down-sampled to hourly frequency for the measurements. For this dataset we reuse an existing implementation and already trained models provided by Zhang and Jiang\footnote{\url{https://github.com/zhykoties/TimeSeries }}. The model is trained on data from 2011/01 to 2014/08 (included) and we perform the attack on test samples from 2014/09.
The input sequence consists of 168 consecutive measurements, and the network predicts the next 24 values (corresponding to the next day). The total number of test samples is 2590.

\paragraph{Experimental Setup}
We performed all experiments on a machine running Ubuntu 18.04, with 2.00GHz Intel Xeon E5-2650 CPU and using a single GeForce RTX 2080 Ti GPU. For the S\&P500 dataset, each model's training time is under one hour, and running the attack on one model for all test-set elements of a period takes approximately 24 hours. For the electricity dataset, running the attack on a batch of 256 test sequences takes approximately three hours.

\subsection{Attacks on Buy/Sell Classification}\label{sec:buy_sell}
We start by considering a classification task on the \sandp{} dataset where each sample is classified as \emph{buy}, \emph{sell} or \emph{uncertain}.
For each stock, we predict the cumulated return using the statistic $\chi(\vy) = y_h/x_{-1} - 1$.
Then, let $\tau$ be a threshold used to decide whether to buy or sell the stock.
Concretely, if the $95\%$ confidence interval (assuming Student's t-distribution) of the estimation $\chi(\vy)$ is entirely above~$\tau$, the stock is classified as buy. If it is entirely below $\tau$, it is classified as sell. Finally, if $\tau$ is inside the confidence interval, the classification is uncertain.
We set $\tau$ to be the average over all stocks of the ground truth cumulated return, which leads to roughly balanced decisions to buy and sell.

We attack the statistic $\chi$ twice. First, we perturb all samples initially classified as buy or uncertain, in order to make it classify as sell. Similarly, we perturb all samples initially classified as sell or uncertain, in order to make it classify as buy. 
The target of the attack is set as $\tau + \lambda$ for the buy attack and $\tau - \lambda$ for the sell attack. We fix $\lambda = 0.03$ in our experiments. This is aimed at making the $95\%$ confidence interval fit entirely in the buy (resp. sell) zone. Indeed, with $10^4$ samples, the interval width order of magnitude is $0.01$. 

The results without a Bayesian observation $z$ are summarized in \Figref{attack} a) and show that the reparametrization estimator is significantly better at generating adversarial examples that the score-function estimator. For example, using $\epsilon=0.1$ the reparametrization estimator attack succeeds in $16\%$ more cases.
The reparametrization estimator can also successfully attack the model when considering a~Bayesian setting with similar results. We include such experiment that uses a smaller horizon $h=5$ and an observation $y_{10}/ x_{-1} = \gamma$ in the supplementary material.

\subsection{Attacks on Long-Short Trading Strategies}
Next, we evaluate the impact of attacking the cumulated return statistic $\chi(\vy) = y_h/x_{-1} -1$ on the financial gain of the long-short trading strategy described in Section~\ref{sec:trading}. We suppose that the attacker is allowed to perturb all inputs at test time without changing the corresponding ground truth outputs, with a maximum tolerance on the perturbation norm. We consider a horizon of $h = 10$ and different portfolio sizes $k$.
Given a ground truth output $\tilde{\vy}$, the target is set to be $t = \tau - \alpha \cdot \left(\chi(\tilde{\vy})-\tau \right)$,
where $\tau$ is the buy/sell threshold defined previously, and $\alpha > 0$ is a scaling factor that rescales the ground truth output to the prediction range of the network.
Intuitively, this attack target corresponds to reversing the prediction of the network around its average, in order to swap the top and flop k stocks.

We report the return of the perturbed portfolios in \Figref{attack}~b).
We observe that the reparametrization estimation is again significantly better compared to the score-function estimator.
Additionally, in both experiments the thresholds for appearance of adversarial perturbation to some of the samples is approximately $\epsilon \approxeq 10^{-2}$.


\subsection{Attacks on Electricity Consumption Prediction}\label{sec:elec}
We perturb each input sequence twice: in order to make the consumption forecast abnormally high (resp. low). We designate these as over-estimation (resp. under-estimation) attack. The attacked statistic is $\chi(\vy) = y_h$, with $h = 18$. Given an input $\vx$, we first approximate the expected value $y^* = \E_{q[\vy|\vx]}[\chi(\vy)]$, and set the target to $t = (1 \pm 0.5) \cdot y^*$ to cause over or under-estimation.

We show the attack success for different perturbation tolerances in \Figref{attack} c). We observe that the reparametrization estimator (continuous line) yields stronger under-estimation of the prediction than the score-function estimator (dashed-line) In \Figref{attack} d), we give examples of generated adversarial samples for the under-estimation attack. We observe a recurrent pattern in the under-estimation attack, where the perturbed prediction matches closely the original prediction for the first time-steps, but eventually becomes significantly inferior. In the supplementary material, we provide similar figures for the over-estimation attack.

%

\ignore{
\subsection{Iterated Stock-Market Prediction}

First, we evaluate the performance of the probabilistic forecasting model compared to state-of-the-art deterministic networks. We consider two baselines as the current state-of-the-art for financial predictions: LSTM networks~\cite{fischer2018deep}, and Temporal Convolutional Networks (TCN)~\cite{borovykh2017conditional}. We describe precisely all the training hyper-parameters in~\ref{app:architectures}.

\subsubsection{Long-Short Trading Strategies}\label{sec:trading}

Given a forecasting horizon $h \in \llbracket 1 , 10 \rrbracket$, we analyze the characteristics of the following portfolio: at time-step $t$, buy (long) the $k$ stocks for which the forecasting network predicts the highest gain, and sell (short) the stocks for which the predicted loss is the highest. This benchmark is a generalization of the one presented in~\cite{fischer2018deep}, where only direct prediction ($h = 1$) is considered. We do not include transaction costs as they apply identically for all networks.

Formally, we consider the statistics $f_x(y) = y_h/x_{241}$ of the output sequence, which corresponds to the gain of investing one dollar in the stock at time $t$, and then selling at time $t + h$. In a non-bayesian setting, we estimate
\begin{equation}\label{obj:trading}
	E_{q_{x}(y)}[f_x(y)]
\end{equation}
via Monte-Carlo sampling for each stock, and buy (resp. sell) the stocks for which the estimation of quantity~\ref{obj:trading} is the highest (resp. the lowest). Note that this setting also applies to the deterministic baselines, it suffices to consider that $q_{x}(y)$ is a Dirac distribution centered in the deterministic prediction. In Figure~\ref{fig:trading}, we summarize the results for different values of $h$ and $k$. We observe that the TCN is consistently outperformed by both DeepAR and the LSTM network. As expected, the performance of DeepAR improves with the number of samples used for Monte-Carlo estimation. With enough samples, DeepAR has similar or better performance than the baselines for the different portfolio sizes and horizons. We note that using more than $10000$ samples does not improve performance. As a result, we stick with this maximal number in the subsequent experiments.

\begin{figure*}
  \includegraphics[width=\linewidth]{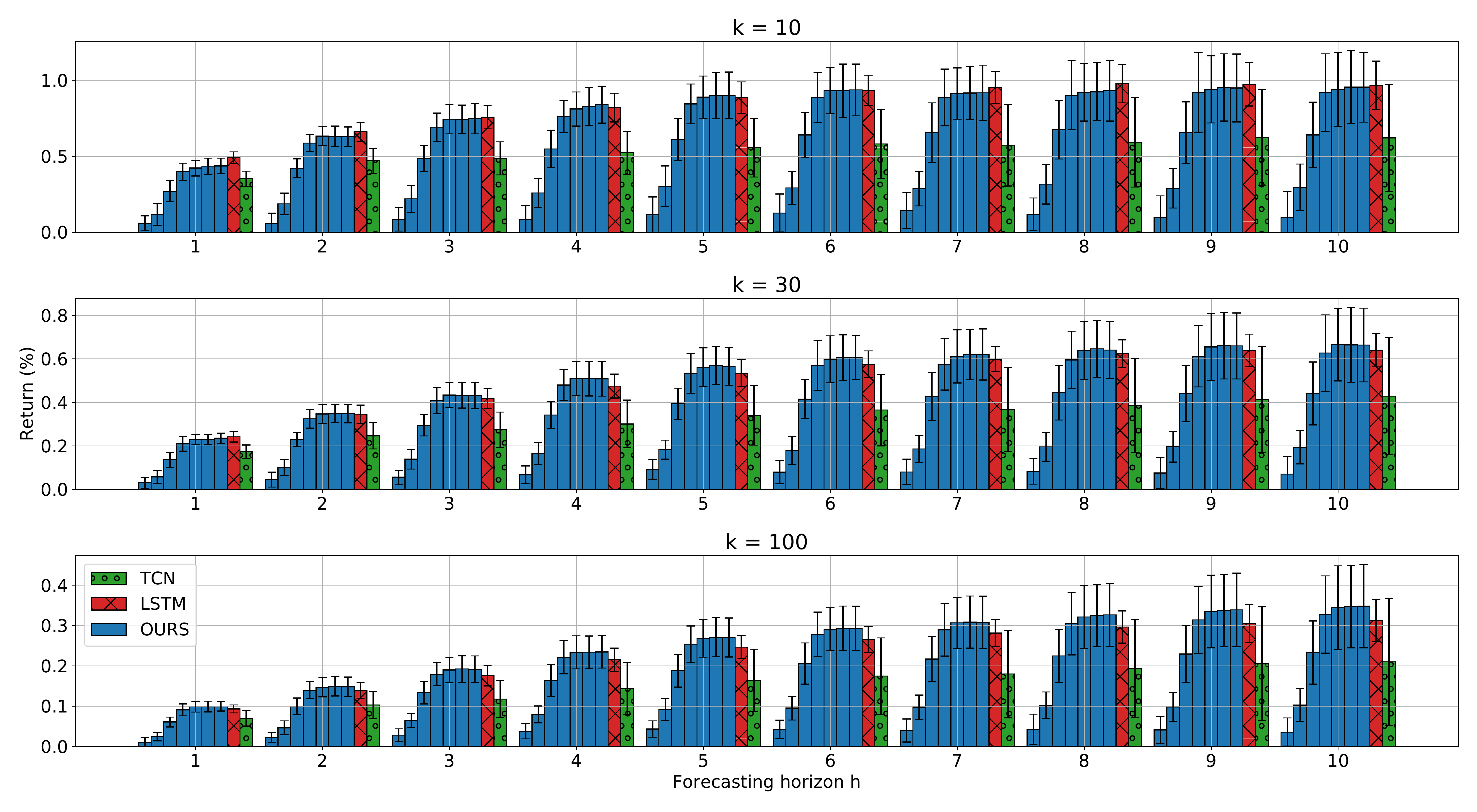}
  \caption{Average portfolio return for different values of the horizon $h$ and the portfolio size $k$. The different blue bars correspond to the number of samples used for Monte-Carlo estimation: from left to right $1, 10, 10^2, 10^3, 10^4, 10^5, 10^6$.}
  \label{fig:trading}
\end{figure*}

\subsubsection{Quality of the Probabilistic Forecast}

We assess the quality of the probabilistic forecast for several statistics of interest in the area of financial market predictions. This includes not only predicting cumulated stock return, but also pricing derivatives such as European call and put options~\cite{black1973pricing}. Besides, as an example of a binary statistic, we propose to predict the probability of success of limit orders~\cite{handa1996limit}. The different statistics are formally defined in Table~\ref{tab:statistics}. To be able to compare deterministic and probabilistic forecast, we use as metric the ranked probability score (RPS)~\cite{weigel2007discrete} of the prediction. This score has the notable advantage that it allows to compare deterministic and probabilistic models. However, it applies only to predictions with finite output space. There exists a continuous version~\cite{gneiting2007strictly}, but it is impractical for our setting because of the memory consumption of computing the score: we favor metrics which can be computed in an on-line fashion with respect to the sampling process. As a result, we first discretize the output before we apply RPS. We present the results in the appendix D. The lower the score, the better the quality of the prediction. Again, we observe that the performance of DeepAR improves with the number of samples, and that it outperforms both baselines provided that enough samples are used.

\begin{table}
\caption{Definition of output sequence statistics for which Ranked Probability Score is measured.}\label{tab:statistics}
\renewcommand{\arraystretch}{1.5}
\center
\begin{tabular}{c | c | c}
Name & Parameters & $f_x(y)$ \\
\midrule
Cum. Return & horizon $h$ &  $\frac{y_h}{x_{241}} $ \\
European Call & horizon $h$, price $p$ & $\max(0,\frac{y_h}{x_{241}}-p)$ \\
European Put & horizon $h$, price $p$ & $\max(0,p-\frac{y_h}{x_{241}})$ \\
Limit Sell & horizon $h$, price $p$ & $\mathbbm{1}_{\frac{\max(y_1,\ldots,y_h)}{x_{241}} \geq p}$ \\
Limit Buy & horizon $h$, price $p$ & $\mathbbm{1}_{\frac{\min(y_1,\ldots,y_h)}{x_{241}} \leq p}$ \\

\end{tabular}
\end{table}
}

\section{Related Work}

\paragraph{Probabilistic Autoregressive Forecasting Models}

Probabilistic autoregressive forecasting models have been used in diverse applications. Schittenkopf et al.~(\citeyear{schittenkopf2000forecasting}) developed the Recurrent Mixture Density Network (RMDN) to predict stock prices volatility iteratively, and were the first to propose using Monte-Carlo methods for iterative prediction. RMDN is based on a vanilla recurrent neural network coupled with Gaussian mixture likelihood. The recent \scode{DeepAR} architecture~\cite{salinas2019deepar} uses several LSTM layers with Gaussian likelihood, and has been applied to forecasting electricity consumption, car traffic and business sales. Follow-up work~\cite{chen2019probabilistic} considers an alternative TCN-based architecture. Our characterization of probabilistic forecasting models encompasses these different architectures, and presents the following novelties \captionn{i} it generalizes inference to any statistic of the output sequence, and \captionn{ii} it extends the prior work to a Bayesian inference setting.

\paragraph{Stock-Market Prediction}
Various methods have been applied for predicting future stock prices including random forests, gradient-boosted trees, or logistic regression. 
Two notable deterministic neural models applied to this task are TCN~\cite{borovykh2017conditional} and LSTM~\cite{fischer2018deep}, which achieved state-of-the-art results.
In~our work, we trained a probabilistic autoregressive model for the same task and achieved comparable or even better results.
Further, there exists a parallel line of work that performs density estimation, but apart from the RMDN~\cite{schittenkopf2000forecasting}, most papers restrict to prediction one-step ahead~\cite{ormoneit1996experiments}, or to less-expressive and solvable dynamics such as the GARCH~\cite{duan1995garch}.

\paragraph{Adversarial Attacks}
A growing body of recent work on adversarial attacks deals with generating small input perturbations causing mis-predictions (see~\cite{wiyatno2019adversarial} for a survey). The objective function defined in Equation~\ref{eq:problem_statement} is standard in generating adversarial examples~\cite{szegedy2013intriguing,carlini2017towards}. However, to the best of our knowledge this is the first time that adversarial attacks are applied to probabilistic autoregressive models. The most related work is the adversarial training of smoothed classifiers~\cite{salman2019provably}, where the noise applied to the input leads to a stochastic behavior. 

\textbf{Robust Algorithms for Financial Decision Making} The adoption of machine learning in financial decision making makes it crucial to develop algorithms robust against small environment variations. Recent work here include robust assessment of loan applications~\cite{ballet2019imperceptible}, deepfakes on accouting journal entries~\cite{schreyer2019adversarial}, robust inverse reinforcement learning on market data~\cite{roa2019adversarial}. Adversarial attacks against stock-market prediction algorithms was studied by ~\cite{fengenhancing}. Compared to the latter, our work is the first to operate on a probabilistic network for iterative prediction.

\textbf{Reparametrization}
The \emph{reparametrization trick} has been applied in several fields under different names: perturbation analysis/pathwise derivatives~\cite{glasserman1991gradient} in stochastic optimization,  stochastic backpropagation~\cite{rezende2014stochastic}, affine independent variational inference~\cite{challis2012affine} or correlated sampling in evaluating differential privacy~\cite{bichsel2018dp}. The actual reparametrization of our model resembles that of the deep generative model of Rezende et al.~(\citeyear{rezende2014stochastic}), but there it is used to perform variational inference.

\section{Conclusion}
In this work, we explored applying adversarial attacks to a~recently proposed probabilistic autoregressive forecasting models.
Our work is motivated by the fact that: \captionn{i} this model has been included in the Amazon SageMaker toolkit and achieved state-of-the-art results on a number of different tasks, and \captionn{ii} adversarial attacks and robustness are pressing and important issues that affect it.

Concretely, we implemented and evaluated two techniques, reparametrization and score-function estimators, that are used to differentiate trough Monte-Carlo estimation inherent to this model and instantiate existing gradient based adversarial attacks.
While we show that both of these techniques can be used to generate adversarial attacks, we evidence that using the reparametrization estimator is crucial for producing adversarial attacks with a small perturbation norm.
Further, we extend the prior work to the Bayesian setting which enables using these models with new types of queries.

\bibliography{paper}

\clearpage
\FloatBarrier
\appendix

We provide the following three appendices:
\begin{itemize}
\item \Appref{proofs} provides proofs of Score-function Estimator~\ref{thm:score_function} and Reparametrization Estimator~\ref{thm:reparametrization_estimator} defined in \secref{attack}.

\item \Appref{datasets} provides details of our datasets, pre-processings steps, architectures and hyper-parameters used in our experiments.

\item \Appref{results} provides extended version of the experiments presented in sections \ref{application} and \ref{eval}.

\end{itemize}

\section{Proofs}\label{proofs}

\begin{nnamedthm}{Score-function Estimator}
In the general Bayesian setting where $\vy \sim q[\cdot|\vx+\vdelta, z]$, the score-function gradient estimator of the expected value of $\chi(\vy)$ is:
\begin{align*}
	& \nabla_\vdelta \EQD \\
	\simeq &\frac{ \sum_{l = 1}^{L} \chi(\vy^l) q[z|\vx+\vdelta, \vy^l]  \nabla_\vdelta \log(q[\vy^l|\vx+\vdelta, z]) }{ \sum_{l = 1}^{L} q[z|\vx+\vdelta, \vy^l] }
\end{align*}
where $\vy^l$ is sampled from the prior distribution $q[\vy|\vx+\vdelta] $, and $q[z|\vx+\vdelta, \vy]$ denotes the probability that $z$ is true knowing that $\vy^l$ is generated.
\end{nnamedthm}

\begin{proof}
The expectation is defined as the following integral over the output space:
\[
\EQD = \int_{\vy} \chi(\vy) q[\vy|\vx+\vdelta,z] dy
\]
Using Leibniz rule, we obtain
\[
\nabla_\vdelta \EQD  = \int_{\vy} \chi(\vy) \nabla_{\vdelta} q[\vy|\vx+\vdelta,z] dy
\]
at every point $\vdelta$ around which the gradient $\nabla_{\vdelta} q[\vy|\vx+\vdelta,z]$ is locally continuous (in the model described in this paper, this regularity condition holds everywhere). The resulting integral can be transformed as follows into an expectation over the distribution $q[\vy|\vx+\vdelta,z]$.
\begin{align*}
&\nabla_\vdelta \EQD \\
= &\int_{\vy} \chi(\vy) \cdot \nabla_{\vdelta} q[\vy|\vx+\vdelta,z] dy \\
= &\int_{\vy} \chi(\vy) q[\vy|\vx+\vdelta,z] \frac{\nabla_{\vdelta} q[\vy|\vx+\vdelta,z]}{q[\vy|\vx+\vdelta,z]}  dy \\
= &\int_{\vy} \chi(\vy) q[\vy|\vx+\vdelta,z] \nabla_\vdelta \log \left(  q[\vy|\vx+\vdelta,z] \right)  dy \\
=\ &\E_{q[\vy|\vx+\vdelta,z]}[\chi(\vy) \nabla_\vdelta \log \left(  q[\vy|\vx+\vdelta,z] \right)]
\end{align*}

This expectation can be approximated via Monte-Carlo methods. While it is in general not possible to directly sample from $q[\vy|\vx+\vdelta,z]$, what can be done instead is generating samples $\vy^l$ for $l \in 1 \leq l \leq L$ from the prior $q[\vy|\vx+\vdelta]$, and attribute an importance weight to each of the resulting samples, yielding:
\begin{align*}
&\nabla_\vdelta \E_{q[\vy|\vx+\vdelta,z]}[\chi(\vy)]\\
 \simeq &\frac{ \sum_{l = 1}^{L} \chi(\vy^l) q[z|\vx+\vdelta, \vy^l]  \nabla_\vdelta \log(q[\vy^l|\vx+\vdelta, z]) }{ \sum_{l = 1}^{L} q[z|\vx+\vdelta, \vy^l] }
\end{align*}
The choice of $q[z|\vx+\vdelta, \vy^l]$ as the importance weight for $\vy^l$ results from the application of Bayes rule:
\[
q[\vy^l|\vx+\vdelta,z] = \frac{q[z|\vx+\vdelta, \vy^l]}{q[z|\vx+\vdelta]} q[\vy^l|\vx+\vdelta]
\]
\end{proof}

\begin{nnamedthm}{Reparametrization Estimator}
Assume there exists a differentiable transformation $g_\vx(\vdelta,\bm{\eta})$ such that the random variable $\vy \sim q[\cdot |\vx+\vdelta] $ can be reparametrized as $\vy = g_\vx(\vdelta,\bm{\eta})$, where $\bm{\eta}$ is an independent random variable whose marginal distribution $p(\bm{\eta})$ is independent from $\vdelta$.  Then the importance sampling reparametrization estimator of the expectation's gradient is:
\begin{align*}
	 &\nabla_\vdelta \E_{q[\vy|\vx+\vdelta, z]}[ \chi(\vy)] \\
	 \simeq &\nabla_\vdelta \left( \frac{ \sum_{l = 1}^L \chi(g_\vx(\vdelta,\bm{\eta}^l))  q[z|\vx+\vdelta, g_\vx(\vdelta,\bm{\eta}^l)]}{ \sum_{l = 1}^L q[z|\vx+\vdelta, g_\vx(\vdelta,\bm{\eta}^l)]  } \right) \\ 
\end{align*}
where for $1 \leq l \leq L$, $\bm{\eta}^l$ is sampled from the distribution $p(\bm{\eta})$, and $\vy^l = g_\vx(\vdelta,\bm{\eta}^l)$.
\end{nnamedthm}

\begin{proof}
Approximating the expectation $ \E_{q[\vy|\vx+\vdelta, z]}[ \chi(\vy)] $ via Monte-Carlo estimation with importance sampling yields:
\begin{align*}
	 &\nabla_\vdelta \E_{q[\vy|\vx+\vdelta, z]}[ \chi(\vy)] \\
	 \simeq &\nabla_\vdelta \left( \frac{ \sum_{l = 1}^L \chi(\vy^l)  q[z|\vx+\vdelta, \vy^l]}{ \sum_{l = 1}^L q[z|\vx+\vdelta, \vy^l]  } \right)
\end{align*}
where $\vy^l$ is sampled from the prior distribution $q[\vy|\vx+\vdelta]$. With the assumptions of the theorem, we can rewrite:
\begin{align*}
 &\left( \frac{ \sum_{l = 1}^L \chi(\vy^l)  q[z|\vx+\vdelta, \vy^l]}{ \sum_{l = 1}^L q[z|\vx+\vdelta, \vy^l]  } \right) \\	
	=   &\left( \frac{ \sum_{l = 1}^L \chi(g_\vx(\vdelta,\bm{\eta}^l))  q[z|\vx+\vdelta, g_\vx(\vdelta,\bm{\eta}^l)]}{ \sum_{l = 1}^L q[z|\vx+\vdelta, g_\vx(\vdelta,\bm{\eta}^l)]  } \right)
\end{align*}
Since the respective effects of the perturbation and of randomness are decoupled in this final expression, it is differentiable with respect to $\vdelta$, which concludes.
\end{proof}

\section{Experimental Details}\label{datasets}

Here we provide details of all our experiments to support reproducibility. Additionally, we will make all our datasets and source code available online.

\subsection{Datasets and Preprocessing}

\paragraph{\sandp 500}
The \sandp 500 dataset is obtained via the \emph{yfinance} API\footnote{\url{https://github.com/ranaroussi/yfinance}}. We focus on data-points between 1990/01 and 2000/12, identified by Fischer and Krauss~(\citeyear{fischer2018deep}) as a period of exceptionally high trading returns compared to the following decades.  We also follow Fischer and Krauss for preprocessing the data. A sequence of prices $ \vp = (p_1,\ldots,p_T)$ is first preprocessed to obtain a sequence of returns $(r_2,\ldots,r_T)$, defined as $r_i = \tfrac{p_{i}}{p_{i-1}} - 1$. Intuitively $r_i$ is the gain (when positive) or loss obtained by investing one dollar in the stock at time $i-1$, and then selling at time $i$. Inversely, given a sequence of returns $\vr$ and an initial price $p_1$, the corresponding sequence of prices can be obtained~as: 

\[\textstyle
p_k = p_1 \prod_{i = 2}^{k} (1 + r_{i})
\]

Both transformations are differentiable, which allows to perform the attack in the application space of prices rather than on returns. Besides, returns are normalized to have zero mean and unit variance. Denoting $\mu$ and $\sigma$ for the mean and standard deviation of returns in the training set, the normalized sequence is $(\tilde{r}_2,\ldots,\tilde{r}_T)$, where $\tilde{r}_i = (r_i - \mu)/\sigma$. We refer to Fischer and Krauss~(\citeyear{fischer2018deep}) for a thorough analysis of the properties of the \sandp 500 dataset.

\textbf{Electricity Dataset}
We use the same preprocessing steps as described in~\cite{salinas2019deepar}. Input sequences are divided by their average value $v$, and the corresponding prediction sequence is multiplied by $v$. This guarantees that all inputs are approximately in the same range.

\subsection{Neural Architectures: \sandp 500 Dataset}

\paragraph{LSTM}

The LSTM baseline used on the \sandp 500 dataset, we follow~\cite{fischer2018deep}, and use a single LSTM layer with $25$ hidden units, followed by a linear output layer. However, we use only one input neuron without activation instead of two neurons with softmax activation.

\textbf{TCN}
In~\cite{borovykh2017conditional}, several sets of hyper-parameters for Temporal Convolutional Networks are used depending on the experiment. We decided to use $8$ layers and a dilation of $2$, in order to match as closely as possible the size of the LSTM receptive field. We selected the other parameters via grid-search, resulting in a kernel size of $2$ and $3$ channels. We use the TCN implementation provided by the authors of~\cite{bai2018empirical}. 

\paragraph{Ours}

For our probabilistic autoregressive model, we chose to use a single LSTM layer with 25 hidden units similar to the LSTM baseline, in order to guarantee the most fair comparison. We only changed the output layers to parametrize a Gaussian distribution. Following~\cite{bishop1994mixture}, we use a linear layer without activation for the mean, and a linear layer with exponential activation for the scale of the distribution. We also performed experiments with a Gaussian mixture likelihood, but it did improve the performance on our two benchmarks.

\paragraph{Training}
For both deterministic networks, we minimize mean-squared error on the training set. For our model, we use negative log-likelihood as a loss function. In both cases, we use the \scode{RMSPROP} optimizer~\cite{tieleman2012lecture} advised by Fischer and Krauss, with default parameters and learning rate of $0.01$. We use an early-stopping patience of $20$, and a large batch size of $2048$ for training. Experiments with different values did not reveal a significant influence of these parameters.

\subsection{Neural Architectures: Electricty Dataset}

\paragraph{\scode{DeepAR}}
The \scode{DeepAR} architecture used for the Electricity experiments is based on a three-layer LSTM with 40 hidden units each. The number of samples used for Monte-Carlo estimation of the output is set to 200. The network is trained for 20 epochs with the \scode{Adam} optimizer~\cite{kingma2014adam}, with batch size of 64 and learning rate of 0.001. 

\begin{figure}
  \includegraphics[width=\linewidth]{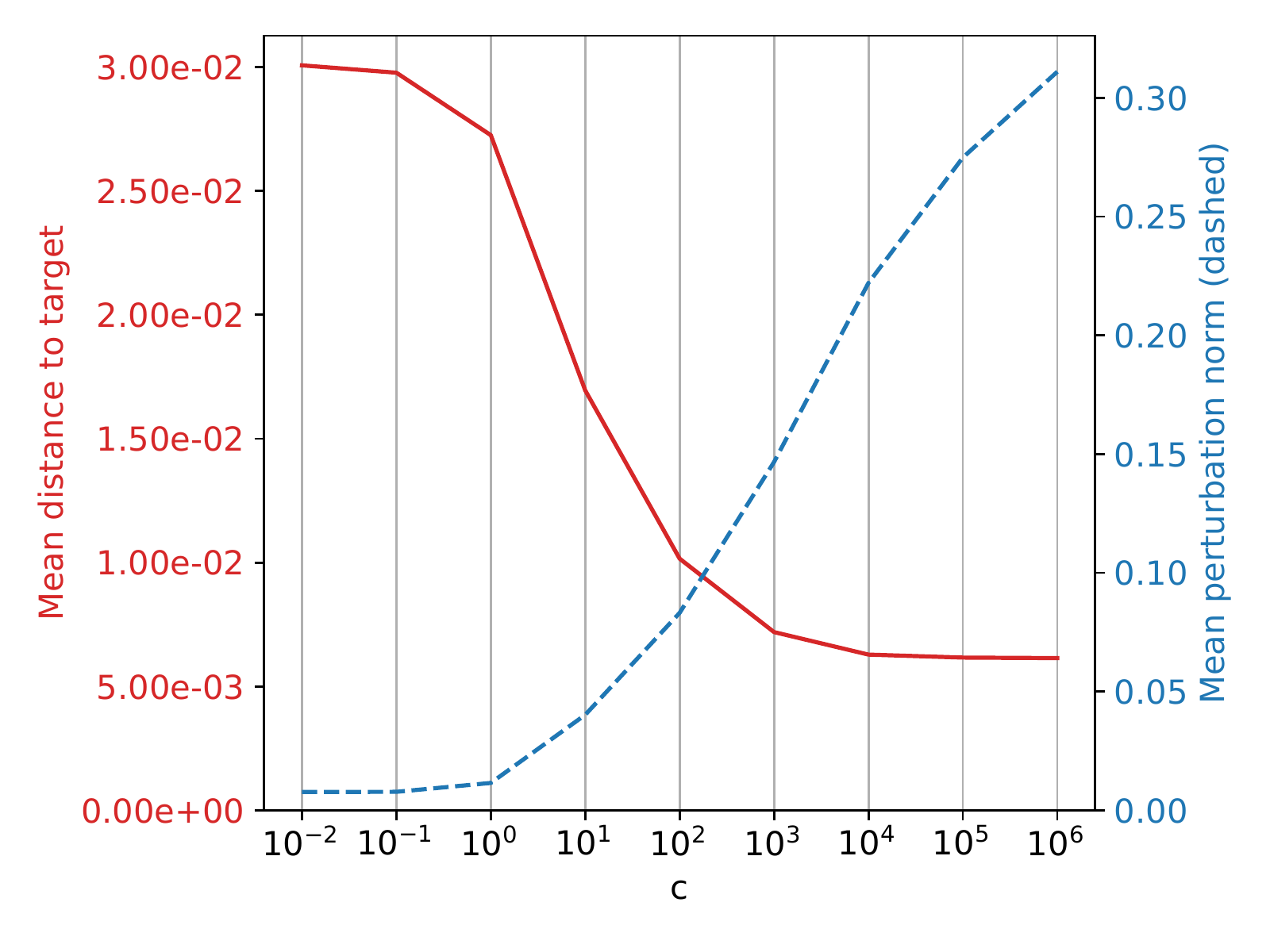}
  \vspace{-2em}
  \caption{Perturbation norm and distance to target for different values of $c$ when evaluated on the \sandp 500 Dataset.}
  \figlabel{param_c}
\end{figure}
\subsection{Attack Hyper-Parameters: \sandp 500 Dataset}


For the \sandp 500 dataset, we optimize the attack objective function with the \textsc{RMSPROP} optimizer using a learning rate of $0.001$ and $1000$ iterations. These parameters were selected with a simple grid search because of the computational cost of running the attack repeatedly. The values used for the coefficient $c$ are $10^{-2}, 10^{-1}, 1, 10, 10^2, 10^3, 10^4, 10^5, 10^6$. We select the value that yields the best adversarial sample under the constraint that the perturbation norm is below the tolerance~$\epsilon$. The number of samples used to estimate the gradient is chosen to be $L = 50$. 

\paragraph{Buy/Sell Attack} We use $\lambda = 0.03$ for the target. For the Bayesian setting, we use $\gamma = y_{10}/x_{-1} = 1.0008$, in order to approximately balance the different classes. The $95\%$ confidence interval is computed assuming Student's t-distribution. In the Bayesian case, the formula for the $95\%$ confidence interval with importance sampling is derived in~\cite{hesterberg1996estimates}.

\paragraph{Attack on Trading Strategies} We use $\alpha = 0.1$ for the target scaling factor.  

\paragraph{Influence of c}

In \Figref{param_c}, we examine the influence of tuning the attack objective function on average perturbation norm and distance to the attack target. We observe a trade-off between these two quantities that depends on the coefficient $c$: higher value for $c$ yields better adversarial samples, at the cost of more input perturbation.

\paragraph{Influence of L}
We evaluate the effect of  the number of samples $L$ used in the reparametrization estimator on the attack loss in \Figref{param_batch_size}.
In this experiment, the value of $c$ is fixed to 1000. We notice a trade-off in terms of convergence speed vs. final loss, that depends on the number of samples used for estimating the gradient. As a result, we choose to use $L = 50$ in our attacks.

\begin{figure}
  \includegraphics[width=\linewidth]{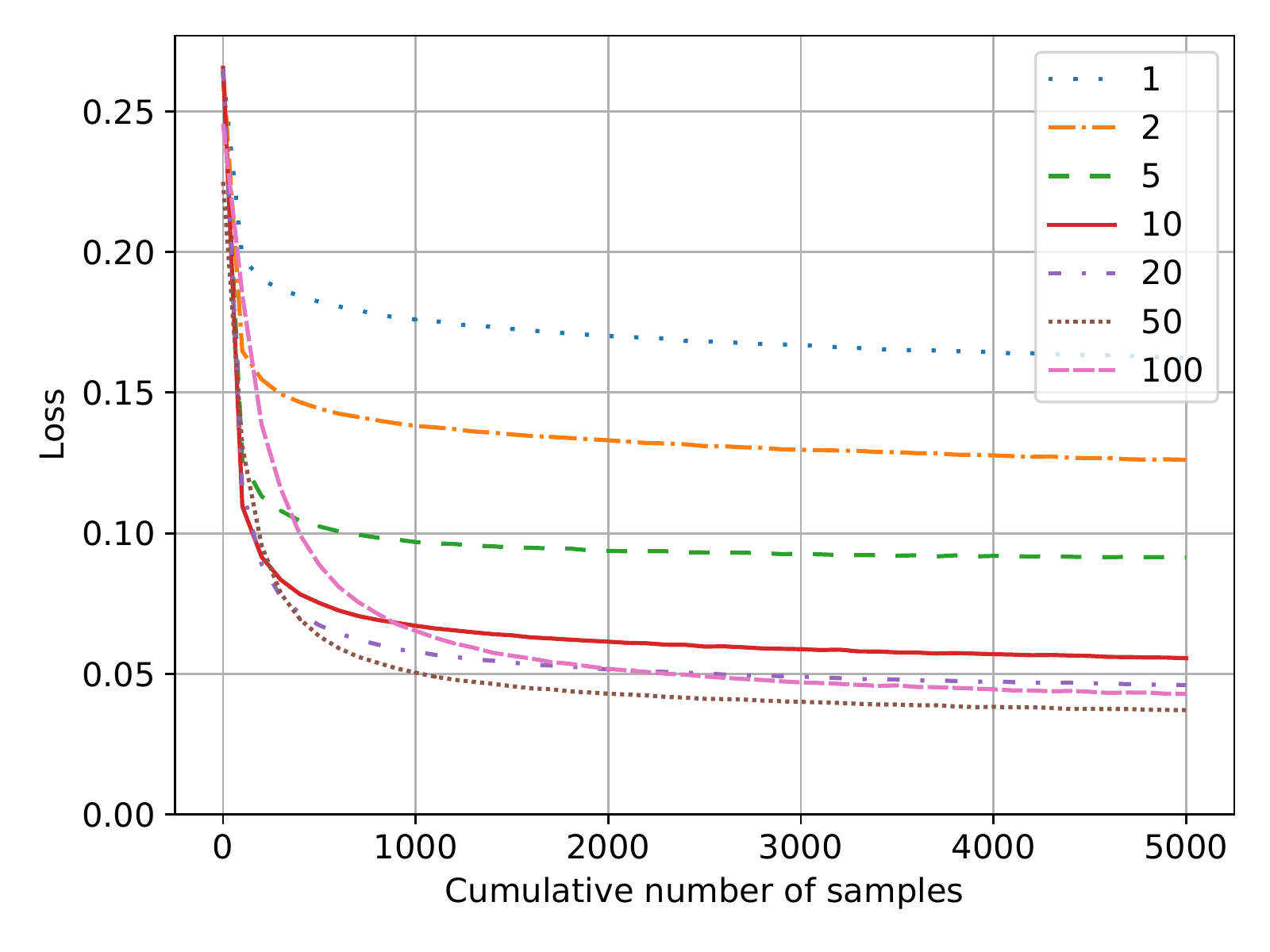}
    \vspace{-2em}
  \caption{Attack loss for different values of $L$ when evaluated on the \sandp 500 Dataset. The x-axis is the total number of generated samples rather than the number of perturbation updates, in order to provide a fair comparison in terms of attack computational cost.}
  \figlabel{param_batch_size}
\end{figure}

\subsection{Attack Hyper-Parameters: Electricity Dataset}

We optimize the attack objective function with the \textsc{Adam} optimizer. We use different optimizers for the two datasets so that the same optimizer is used for training the network and to attack it. We use a learning rate of $0.01$ and $1000$ iterations. These parameters were also selected via informal search. The values used for the coefficient $c$ are 0.1, 0.2, 0.3, 0.5, 0.7, 1, 2, 3, 5, 7, 10, 20, 30, 50, 70, 100, 200 and 300. The number of samples used to estimate the gradient is chosen to be $L = 50$.

\begin{figure*}
  \includegraphics[width=\linewidth]{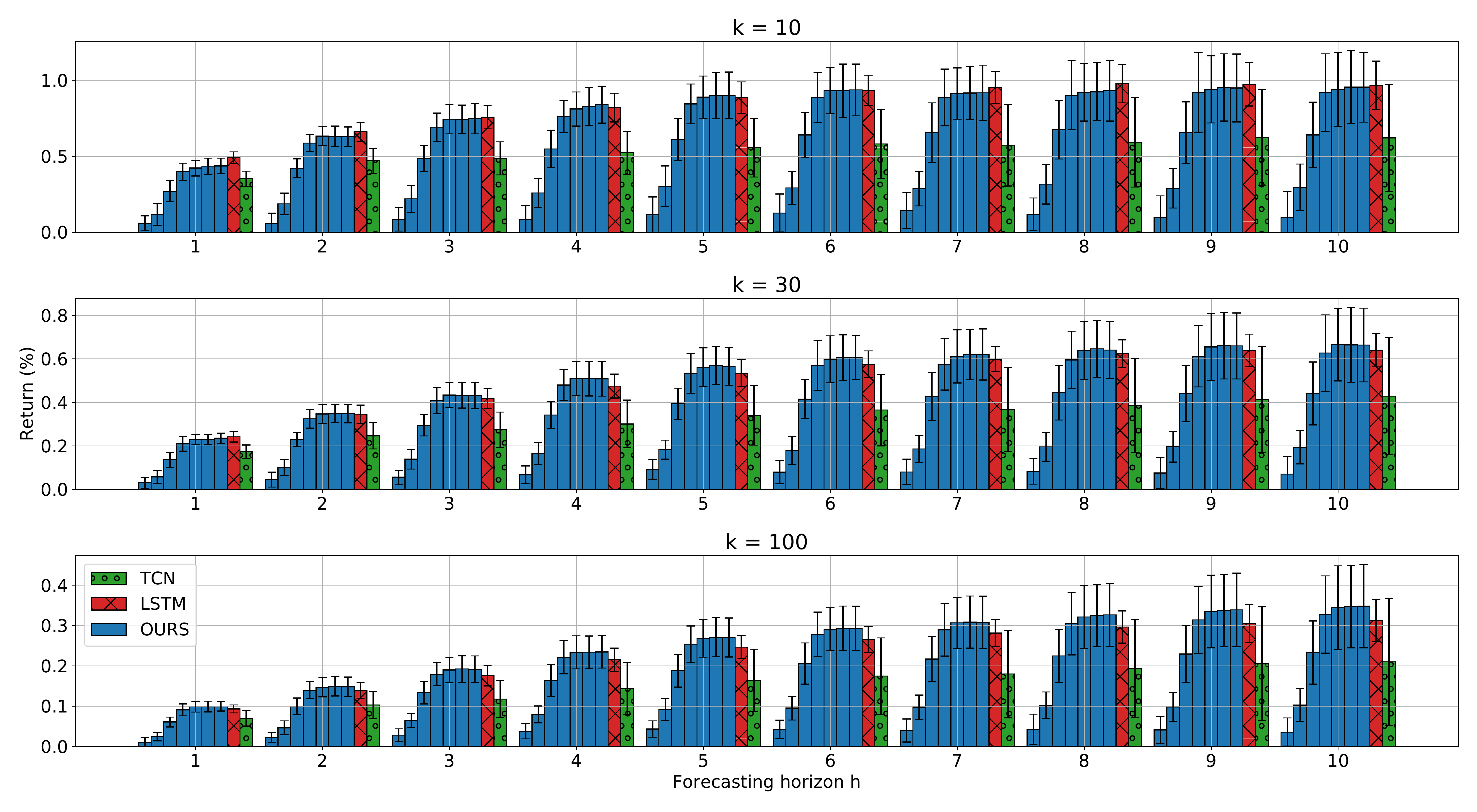}
  \vspace{-2em}
  \caption{Financial gain on algorithmic trading tasks for different horizons $h$ and portfolio sizes $k$ (in $\%$ of the invested capital). The blue bars correspond to different number of samples for Monte-Carlo estimation of the prediction: from left to right $1, 10, 10^2, 10^3, 10^4, 10^5, 10^6$.}
  \label{fig:trading}
\end{figure*}

\section{Experimental Results}\label{results}

\subsection{Trading Strategies}

In Figure~\ref{fig:trading}, we provide extended results for the long-short trading benchmark (Section~\ref{sec:performance}), with different horizons~$h$ and number of samples used for Monte-Carlo estimation of the prediction. We observe that the quality of the probabilistic prediction improves with the number of samples until $10^4$ samples. Further increasing the number of samples does not yield significant performance improvements.

\subsection{Evaluation of the Probabilistic Forecast}

In Table~\ref{rps_detailed}, we give detailed results for the comparison of probabilistic forecasts quality with Ranked Probability Skill (a summary of these results is provided in Table 2, Section~\ref{sec:performance}). We observe that the forecasting quality of our model improves with the number of samples, and that an order of magnitude of the number of samples needed to obtain the best possible estimation is $10^4$. As a comparison, the \scode{DeepAR} implementation on the electricity dataset uses 200 samples. We surmise that this discrepancy is due to the low signal-to-noise ratio of financial data, that makes inference more difficult.

\subsection{Bayesian Attack}

\begin{figure}[t]
  \includegraphics[width=\linewidth]{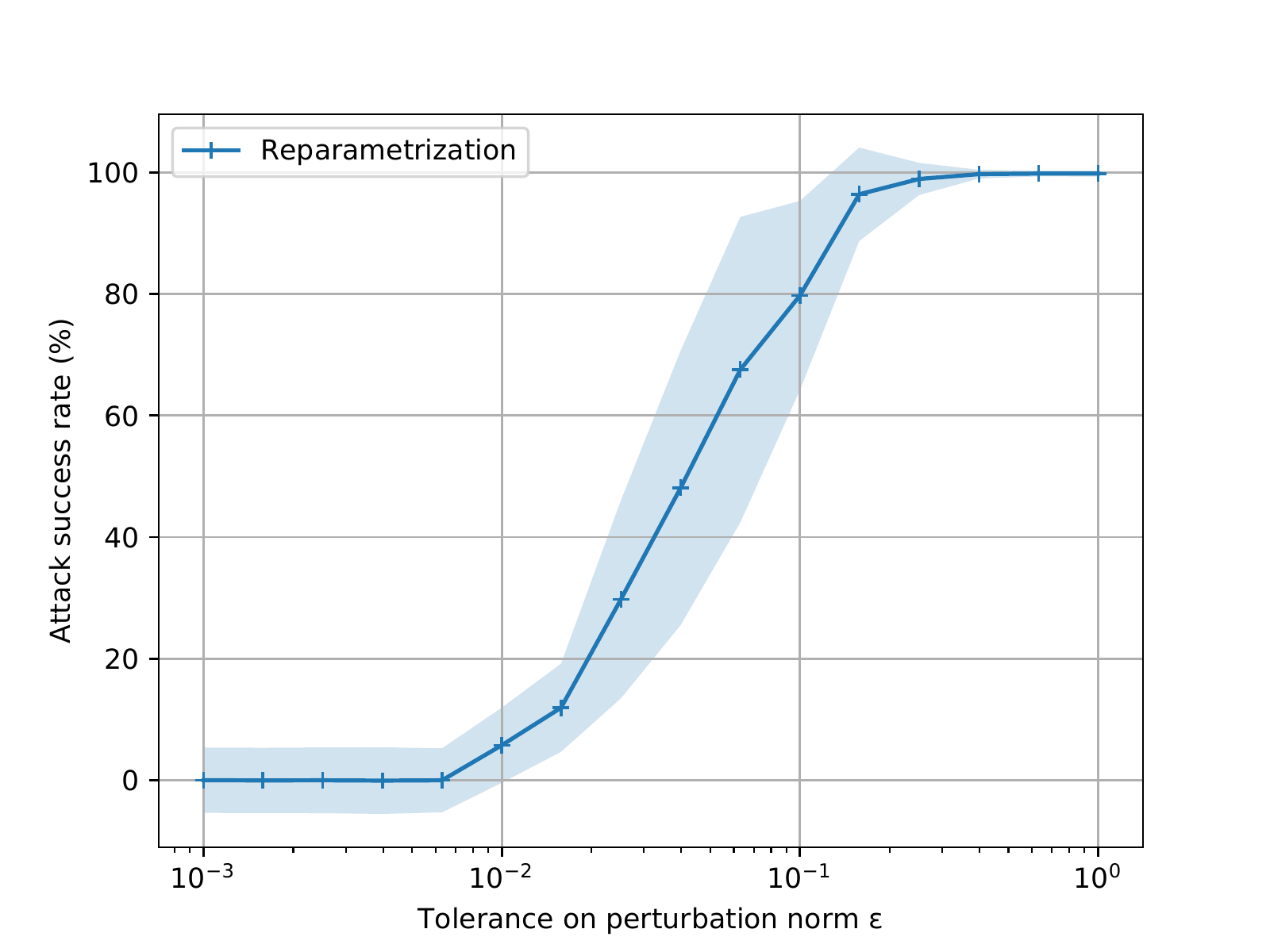}
  \vspace{-2em}
  \caption{Success rate of the classification attack for different perturbation norms, in a Bayesian setting with observation $y_{10}/x_{-1} = \gamma = 1.0008$, and prediction horizon $h = 5$.}
  \label{fig:Bayesian}
\end{figure}

In Figure~\ref{fig:Bayesian}, we plot the results of the classification attack in the Bayesian setting with observation $y_{10}/x_{-1} = \gamma$, where $\gamma = 1.0008$ (Section~\ref{sec:buy_sell}). We only implemented the reparametrization estimator, as the score-function estimator requires the overly complex estimation of $\nabla_\vdelta \log(q[ \vy^l|\vx+\vdelta,z])$ for each sample $\vy^l$. We observe that the attack success rate is very similar to the non-Bayesian setting, demonstrating that the reparametrization estimator adapts readily to the Bayesian setting. The attack success rate is approximately $80\%$ for $\epsilon = 0.1$. 

\subsection{Electricity Dataset}
\begin{figure*}[hpb]
  \includegraphics[width=\linewidth]{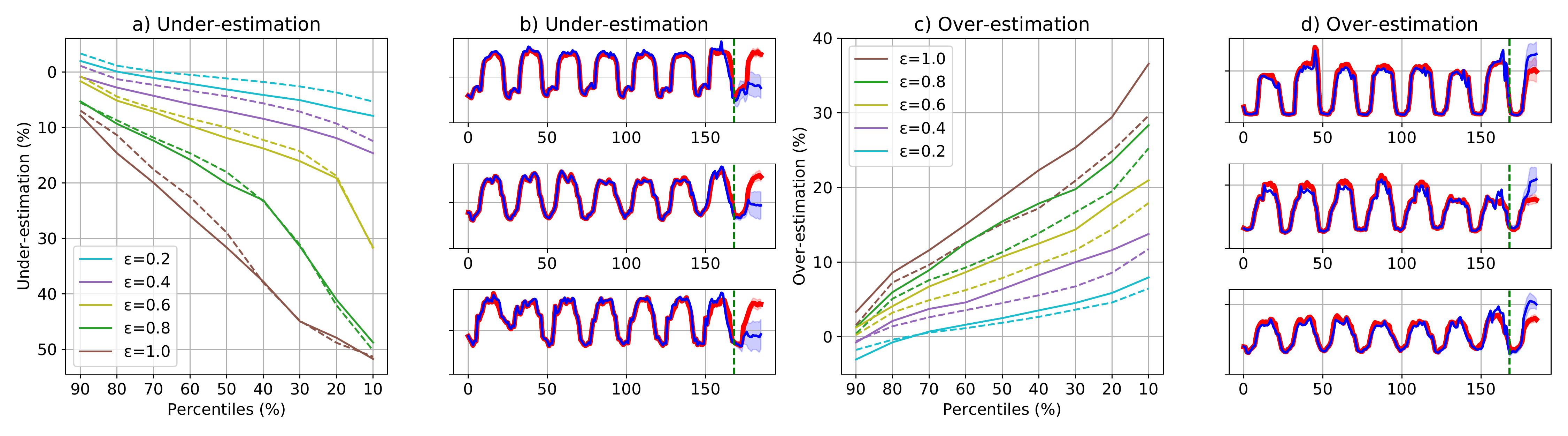}
  \vspace{-1em}
  \caption{Results for the electricity dataset. a) Under-estimation of electricity consumption. For example, with $\epsilon = 1.0$, the attack using reparametrization estimator leads to under-estimation of at least $20\%$ (y-axis) for $70\%$ of samples (x-axis). b)~Under-estimation adversarial samples. c) Over-estimation of electricity consumption. d) Over-estimation adversarial samples. In a) and b), results are given for reparametrization (continuous) and score-function (dashed) estimators. In c) and d), the reparametrization estimator is used, and $\epsilon$ is fixed to 0.9. Red curve is the original sample,
blue curve is the generated adversarial sample. The vertical dashed line separates the input sequence from the network's prediction. }
  \label{fig:elec}
\end{figure*}

In Figure~\ref{fig:elec}, we show results of both over-estimation and under-estimation attacks on the electricity dataset, with examples of generated adversarial samples (Section~\ref{sec:elec}). We observe that for equal perturbation tolerance, the over-estimation attack yields mis-predictions of smaller amplitude. For instance, the reparametrization attack with $\epsilon = 0.8$ causes median over-estimation of around $15\%$, whereas it causes median under-estimation of $20\%$. We believe that this is due to the particular nature of the dataset rather than asymmetry in the attack. We do not observe such a discrepancy in the financial experiments.

\begin{table*}[t]
\caption{Performance of different models on probabilistic forecast of various statistics. The comparison metric is Ranked Probability Skill~\cite{weigel2007discrete} of the prediction (lower scores correspond to better predictions). The performance of our architecture is given for different number of samples used in the Monte-Carlo estimation (1, 100, and 10000).}\label{rps_detailed}
\tra{1.3}
\center
\addtolength{\tabcolsep}{-2pt}
\begin{tabular}{p{16mm} c c c c c c c c c c c c}
\toprule
\multicolumn{3}{c}{Statistics} & &  \multicolumn{2}{c}{Non-probabilistic} & & \multicolumn{3}{c}{Probabilistic} \\ 
\cline{1-3} \cline{5-6}\cline{8-10} 
Name  &  $h$ & $\pi$ &  &  \scode{TCN} & \scode{LSTM}  & & \multicolumn{3}{c}{Ours} \\[-0.7em]
& & & &  \scriptsize \cite{borovykh2017conditional} & \scriptsize \cite{fischer2018deep} & & 1 sample & 100 samples & 10000 samples \\
\midrule
\multirow{3}{*}{\shortstack{Cumulated\\ Return}}  & 1 & - & &  1.423 ($\pm$ 0.022) &  1.424 ($\pm$ 0.016) & & 2.016 ($\pm$ 0.023) &  0.992 ($\pm$ 0.002) &  \textbf{0.982 ($\pm$ 0.002)} \\
	 & 5 & -  & &1.468 ($\pm$ 0.01) &  1.466 ($\pm$ 0.008) & & 1.992 ($\pm$ 0.013) &    1.0 ($\pm$ 0.005) &   \textbf{0.99 ($\pm$ 0.004)} \\
	 & 10 & - & &  1.548 ($\pm$ 0.029) &  1.541 ($\pm$ 0.019) &  & 1.995 ($\pm$ 0.011) &  1.012 ($\pm$ 0.008) &  \textbf{1.002 ($\pm$ 0.008)} \\
	 \midrule
\multirow{3}{*}{\shortstack{European\\ Call Option}} 	  & 10 & 0.9 & & 1.019 ($\pm$ 0.004) &  1.017 ($\pm$ 0.003) & &  1.961 ($\pm$ 0.22) &  0.999 ($\pm$ 0.009) &  \textbf{0.989 ($\pm$ 0.009)} \\
 	  &10  & 1 & & 1.122 ($\pm$ 0.002) &  1.121 ($\pm$ 0.002) & & 1.966 ($\pm$ 0.103) &  0.992 ($\pm$ 0.006) &  \textbf{0.982 ($\pm$ 0.005)} \\
 	  & 10 & 1.1 &  & 1.342 ($\pm$ 0.002) &  1.341 ($\pm$ 0.002) & &  1.987 ($\pm$ 0.03) &  1.003 ($\pm$ 0.007) &  \textbf{0.993 ($\pm$ 0.007)} \\
 	  \midrule
\multirow{3}{*}{\shortstack{European\\ Put Option}}	  & 10 & 0.9 & & 1.445 ($\pm$ 0.021) &  1.445 ($\pm$ 0.017) & & 1.984 ($\pm$ 0.015) &  1.002 ($\pm$ 0.007) &  \textbf{0.992 ($\pm$ 0.007)} \\
 	  & 10 & 1 & & 1.302 ($\pm$ 0.003) &    1.3 ($\pm$ 0.002) & & 1.957 ($\pm$ 0.036) &  0.984 ($\pm$ 0.005) &  \textbf{0.974 ($\pm$ 0.005)} \\
 	  & 10 & 1.1 &  & 1.046 ($\pm$ 0.005) &  1.044 ($\pm$ 0.002) & & 1.856 ($\pm$ 0.094) &  0.968 ($\pm$ 0.004) &  \textbf{0.959 ($\pm$ 0.003)} \\
\midrule
\multirow{3}{*}{\shortstack{Limit Sell}} 		  & 10 & 1.01 & & 2.822 ($\pm$ 0.501) &  3.137 ($\pm$ 0.307) & & 1.917 ($\pm$ 0.021) &  1.013 ($\pm$ 0.008) &  \textbf{1.004 ($\pm$ 0.008)} \\
 		 & 10 & 1.05 & & 1.516 ($\pm$ 0.001) &  1.514 ($\pm$ 0.002) & & 1.899 ($\pm$ 0.027) &  0.953 ($\pm$ 0.006) &  \textbf{0.944 ($\pm$ 0.006)} \\
 		  & 10 & 1.20 & & 1.035 ($\pm$ 0.003) &  1.034 ($\pm$ 0.002) & &  1.792 ($\pm$ 0.12) &  0.951 ($\pm$ 0.006) &  \textbf{0.942 ($\pm$ 0.005)} \\
 		  \midrule
\multirow{3}{*}{\shortstack{Limit Buy}} 		  & 10 & 0.8 & & 1.02 ($\pm$ 0.002) &  1.019 ($\pm$ 0.002) & & 1.948 ($\pm$ 0.223) &  0.982 ($\pm$ 0.015) &  \textbf{0.972 ($\pm$ 0.013)} \\
  & 10 & 0.95 & & 1.412 ($\pm$ 0.0) &   1.41 ($\pm$ 0.001) & & 1.926 ($\pm$ 0.039) &  0.967 ($\pm$ 0.009) &  \textbf{0.958 ($\pm$ 0.008)} \\
  & 10 & 0.99 & & 3.047 ($\pm$ 0.012) &  3.025 ($\pm$ 0.028) & & 1.963 ($\pm$ 0.024) &  1.013 ($\pm$ 0.006) &  \textbf{1.003 ($\pm$ 0.006)} \\
\bottomrule
\end{tabular}
\end{table*}

\end{document}